\documentclass[letterpaper, 10 pt, conference]{ieeeconf}  

\IEEEoverridecommandlockouts                              

\usepackage{graphics} 
\usepackage{epsfig} 
\usepackage{mathptmx} 
\usepackage{times} 
\usepackage{amsmath} 
\usepackage{amssymb}  
\usepackage{comment}
\usepackage[T1]{fontenc}
\usepackage{multirow}
\usepackage{textcomp}
\usepackage{gensymb}
\usepackage{balance}
\usepackage{soul}
\usepackage{color}

\usepackage[colorinlistoftodos]{todonotes} 

\usepackage{lipsum}
\usepackage{graphicx}
\usepackage[caption=false, font=footnotesize]{subfig}

\usepackage[font=small, belowskip=-10pt, aboveskip=5pt]{caption}
\usepackage{enumerate}

\usepackage{adjustbox}
\usepackage[utf8]{inputenc}
\usepackage{csquotes}
\usepackage[english]{babel}

\usepackage[backend=biber,style=numeric,sorting=none, maxnames=99, giveninits=true]{biblatex} 
\usepackage{titlesec}

\titlespacing\section{0pt}{5pt plus 4pt minus 2pt}{0pt plus 2pt minus 2pt}
\titlespacing\subsection{0pt}{5pt plus 4pt minus 2pt}{0pt plus 2pt minus 2pt}
\titleformat{\subsubsection}[runin]{}{}{}{\itshape}[]

\title{\LARGE \bf
Autonomous Robotic Mapping of Fragile Geologic Features
}

\author{Zhiang Chen, J Ram\'{o}n Arrowsmith, Jnaneshwar Das
\thanks{Authors are affiliated with the School of Earth and Space Exploration, Arizona State University, 781 Terrace Mall, Tempe, AZ 85287, USA
        }%
}

\begin{document}

\maketitle
\thispagestyle{empty}
\pagestyle{empty}
\maxdeadcycles=200

\begin{abstract}
Robotic mapping is useful in scientific applications that involve surveying unstructured environments. This paper presents a target-oriented mapping system for sparsely distributed geologic surface features, such as precariously balanced rocks (PBRs), whose geometric fragility parameters can provide valuable information on earthquake shaking history and landscape development for a region. With this geomorphology problem as the test domain, we demonstrate a pipeline for detecting, localizing, and precisely mapping fragile geologic features distributed on a landscape. To do so, we first carry out a lawn-mower search pattern in the survey region from a high elevation using an Unpiloted Aerial Vehicle (UAV). Once a potential PBR target is detected by a deep neural network, we track the bounding box in the image frames using a real-time tracking algorithm. The location and occupancy of the target in world coordinates are estimated using a sampling-based filtering algorithm, where a set of 3D points are re-sampled after weighting by the tracked bounding boxes from different camera perspectives. The converged 3D points provide a prior on 3D bounding shape of a target, which is used for UAV path planning to closely and completely map the target with Simultaneous Localization and Mapping (SLAM). After target mapping, the UAV resumes the lawn-mower search pattern to find the next target. We introduce techniques to make the target mapping robust to false positive and missing detection from the neural network. Our target-oriented mapping system has the advantages of reducing map storage and emphasizing complete visible surface features on specified targets. 
\end{abstract}

\section{Introduction}
\begin{figure}
\centering
\includegraphics[width=0.45\textwidth]{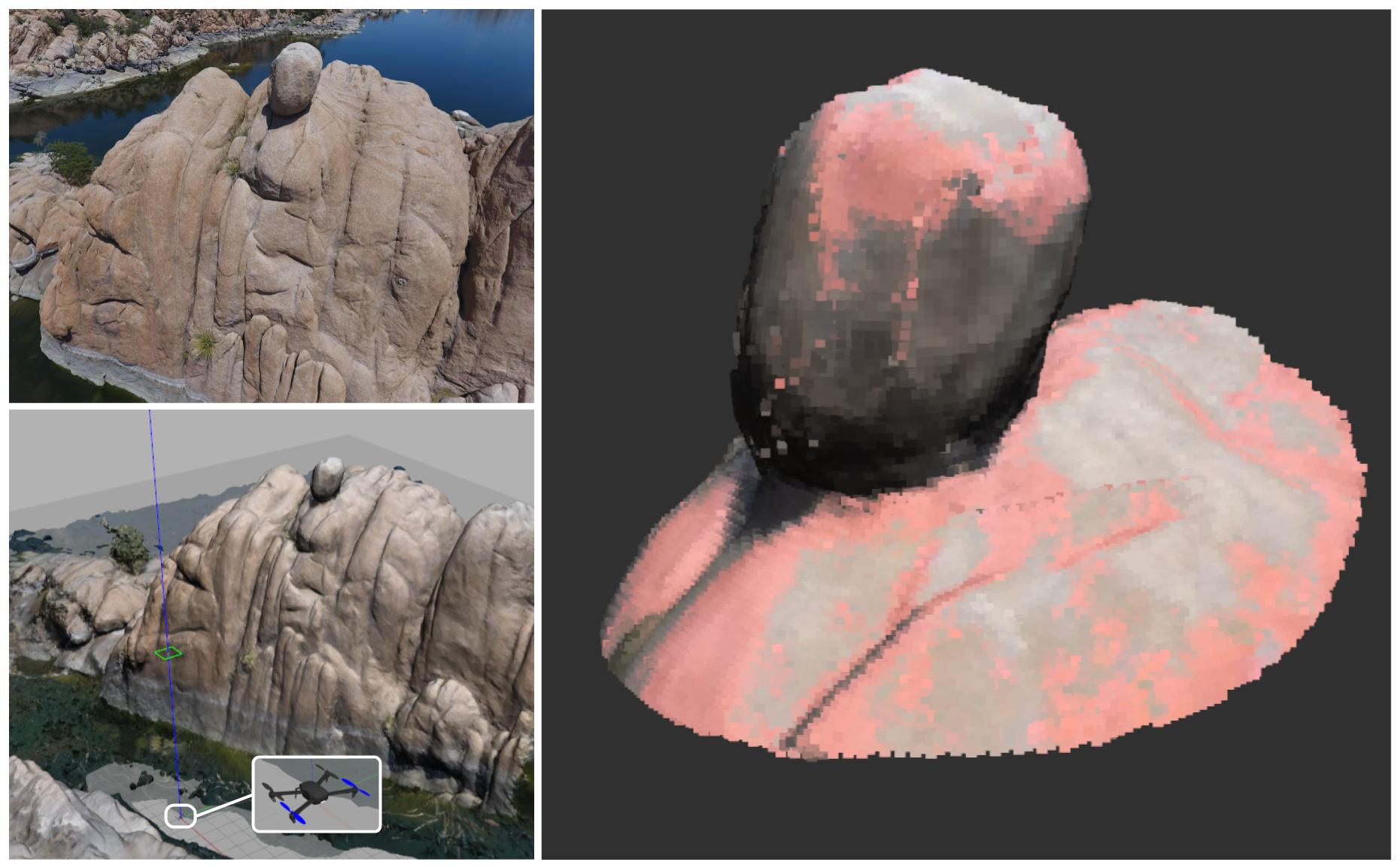}
\caption{(Top left) A precariously balanced rock (PBR) in Granite Dells, Arizona. (Bottom left) Reconstructed world in Gazebo. (Right) Resulting mapped PBR from the reconstructed scene.}
\label{fig:granite_dell}
\end{figure}

This work is motivated by the challenges of finding and mapping fragile geologic features such as precariously balanced rocks (PBRs) (Fig.~\ref{fig:granite_dell} (top left)). PBRs are fragile to being easily toppled by strong earthquake shaking, so their existence can provide valuable information about earthquake history (or lack thereof) in a given region~\cite{brune1996precariously, anooshehpoor2004methodology}. While current studies of such fragile geologic features have focused on single PBR models~\cite{haddad2012estimating}, large datasets of spatially explicit PBR fragility with detailed geomorphic and geologic context provide a valuable assessment of sensitivity to ground motions. Instead of having a few fragilities, it should be possible to have a fragility histogram for the PBRs at a site. However, model assessment in a broad statistical sense is challenging because PRBs are sparsely distributed and difficult to inventory. Additionally, it is critical to reconstruct PBR base since ground contact points play a significant role in characterizing fragilities~\cite{wittich20183}. Our ultimate goal is to assess the precarious features broadly, increasing the number of analyses and accounting for their geomorphic setting~\cite{anooshehpoor2004methodology, haddad2012applications}. 

Mapping methods such as Structure from Motion~\cite{johnson2014rapid, chen2019geomorphological}, Visual-SLAM~\cite{mur2017orb, labbe2019rtab}, and LiDAR SLAM~\cite{droeschel2018efficient}, produce dense terrain maps that include both targets of interest and other irrelevant features. As the sparsity of targets increases, significant amounts of computation and memory storage are spent on reconstructing irrelevant features. Current PBR mapping work~\cite{wittich20183} bypasses this problem by focusing on the reconstruction of single PBRs that have been found by geologists. More importantly, mapping methods alone decouple robot navigation from mapping, which inhibits adaptive flight paths from emphasizing the mapping of complete visible surface features on specified targets. 

Robotic search and rescue systems provide approaches to target localization~\cite{tomic2012toward}, however, knowing target location (i.e., geometric center) alone is insufficient for target mapping due to ambiguity in target contours, which might cause collision problems if the visual-motion system is not carefully adapted. Recent object-level SLAM~\cite{mccormac2018fusion++} is appealing for target mapping but limited to room-scale environments because the target tracking and localization depend on a depth camera, whose sensing range is inadequate for field applications. Other object-level SLAM based on monocular cameras~\cite{nicholson2018quadricslam} approximates objects with ellipsoids, which convey target pose and occupancy information. However, to precisely map a target, additional processes are still required. Additionally, object-level SLAM algorithms have not explicitly considered false positive detection from neural networks. False positive detection is less frequent when neural networks are trained from benchmark datasets, but has commonly been seen in domain-specified, practical applications.

\begin{figure}
\centering
\vspace{6pt}
\includegraphics[width=0.48\textwidth]{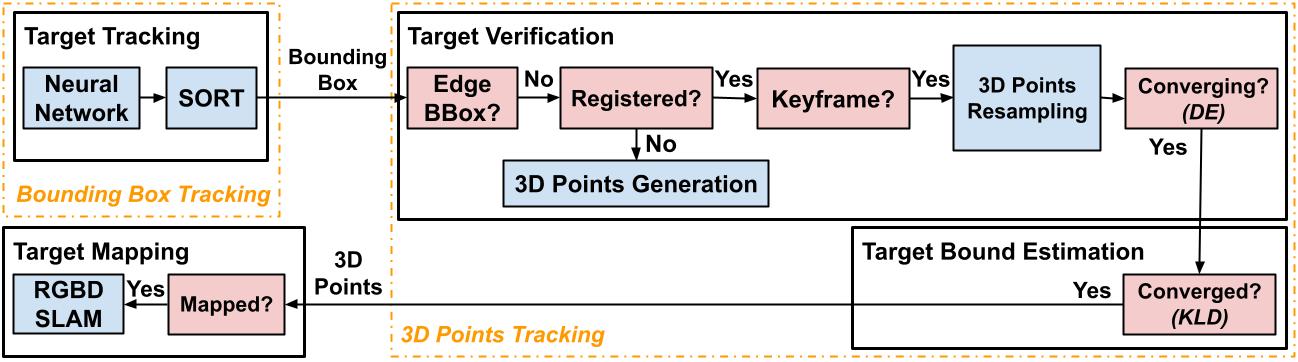}
\caption{Perception subsystem workflow.}
\label{fig:perception_workflow}
\end{figure}

This paper presents an efficient, real-time, and robust system to detect, localize, and map fragile geologic features, i.e. PBRs, in a landscape. The autonomous system for PBR search and mapping provides access to fragility distribution, which is valuable to earthquake studies but has not been obtained due to the challenge of collecting large datasets. By introducing two tracking modules, the proposed target-oriented mapping system is robust to false positive or missing detection from a deep neural network. We implement the system using Robot Operating System with the PX4 UAV flight stack\footnotemark. 

\footnotetext{\url{https://github.com/ZhiangChen/target_mapping}}

\section{Problem Description}
The goal is to search for and map PBRs in a given survey region. The system should only map the targets of interest and also emphasize complete visible surface features of the targets including visible contact points with the ground. The hardware platform considered within this work is a UAV equipped with a flight controller, one/two onboard companion computers, a GPS module, and a stereo camera. The onboard companion computers, e.g. Intel NUC i7 + Nvidia TX2 in practice, should allow a lightweight detection neural network to conduct real-time inference. There are bounded noises in UAV's pose estimation, which is feasible for a flight controller (e.g. Pixhawk with PX4 flight stack) running EKF with GPS and IMU. A stereo camera is used for target search and mapping. The UAV's relative elevation with respect to the terrain during search can be greater than the depth range of the stereo camera, so the target search only relies on one monocular camera from the stereo camera, and the close and detailed mapping uses the stereo camera to generate dense point clouds. 

We assume there are no obstacles around PBRs during UAV mapping. The local terrains of PBRs are within a certain slope $\Psi$ such that the local terrains cannot present obstacles for UAV mapping. There might be false positive or missing detections from neural networks, so the system should be robust to such false detections. The UAV camera can capture the full body of each PBR during search, which is feasible in practice by adjusting the elevation of UAV search pattern. 

\section{System Overview}
\begin{figure} 
\vspace{6pt}
    \centering
  \subfloat[Motion state diagram \label{motion_state}]{%
       \includegraphics[width=0.65\linewidth]{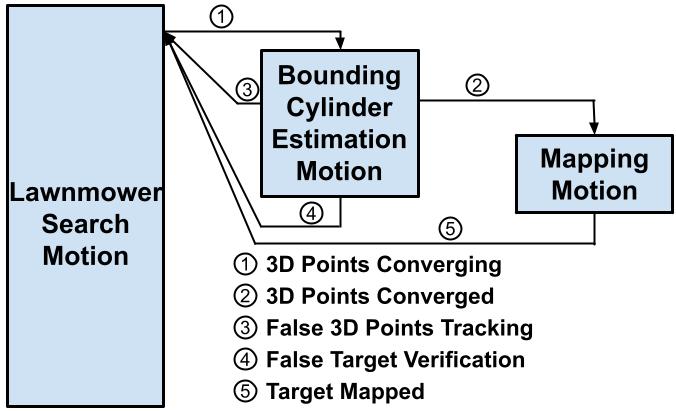}}
    \\
  \vspace{-5pt}
  \subfloat[Controller workflow \label{control_system}]{%
        \includegraphics[width=0.8\linewidth]{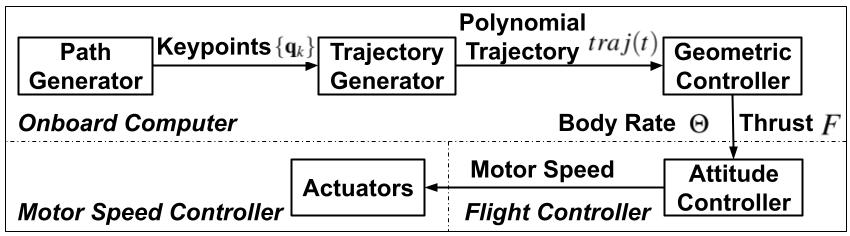}}
  \caption{Motion subsystem\protect.}
  \label{fig:motion_system} 
\vspace{-6pt}
\end{figure}

The proposed target-oriented mapping system is composed of a perception subsystem (Fig.~\ref{fig:perception_workflow}) and a motion subsystem (Fig.~\ref{fig:motion_system}). We start by carrying out a lawn-mower search pattern in the survey region from a high elevation using a UAV. Once a potential PBR target is detected by a deep neural network, we track the target bounding box in image coordinates using a bounding box tracking algorithm. We generate 3D points from the tracked bounding box and resample them with our sampling-based filtering algorithm. When the 3D points are assessed to be converging, the UAV employs a circular motion at the same search elevation to keep filtering 3D points. A bounding cylinder (b-cylinder) is then constructed from the converged 3D points for UAV path planning to closely and completely map the target with SLAM. After target mapping, the UAV resumes the lawn-mower search pattern to find the next target.

\section{Perception Subsystem}
The perception subsystem workflow is shown in Fig.~\ref{fig:perception_workflow}. We deploy two tracking modules: one for bounding boxes and one for 3D points. The bounding box tracking module subscribes to detection messages from a neural network and publishes tracked bounding boxes to the 3D points tracking module. The essence of the 3D points tracking module is a sampling-based filtering algorithm where a set of 3D points are re-sampled after weighting by the tracked bounding boxes from different camera perspectives (see Sec. IV B). We construct a b-cylinder from the converged 3D points and conduct close and complete target mapping with SLAM. 

\subsection{Bounding Box Tracking}
We apply the simple online and real-time tracking (SORT) algorithm~\cite{bewley2016simple} to track bounding boxes from a lightweight detection neural network. When a new detection is associated with a tracked bounding box using intersection-over-union (IoU) and Hungarian algorithm, SORT algorithm updates the tracked bounding box state by maintaining a Kalman filter. 

When a PBR enters the image, a unique identity is registered once the PBR has been detected by the neural network for $T_{bbox\_hit}$ frames. This can prevent false positive detections in a certain sense. A tracked bounding box is deregistered if it has not been continuously detected from $T_{bbox\_missing}$ frames, which can alleviate missing detection issues. Only registered bounding boxes are published to the 3D points tracker. 

\subsection{3D Points Tracking}
\begin{figure}
\centering
\vspace{6pt}
\includegraphics[width=0.40\textwidth]{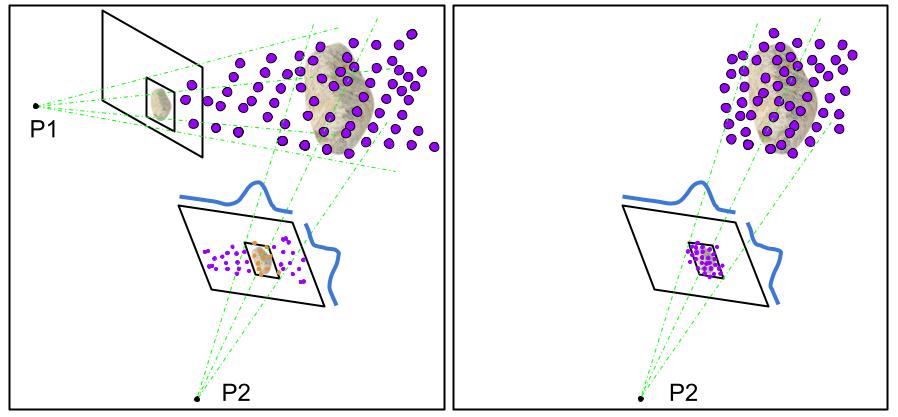}
\caption{3D points generation and update. (Left) 3D points are generated within a polyhedral cone that is back-projected from an enlarged bounding box on image frame P1; 3D points are projected to a different image frame P2 and resampled according to the positions of their 2D projection points with respect to the new bounding box. (Right) Resampling results.}
\label{fig:depth_filter}
\end{figure}

Having received a new tracked bounding box, the 3D points tracker first examines if the bounding box is on the image edge. We drop the ones on the image edge to ensure the bounding box captures the entire body of a target. Then a set of 3D points are generated within a polyhedral cone that is back-projected from the tracked bounding box (Fig.~\ref{fig:depth_filter}). The 3D points will be projected to different image frames and re-sampled according to the positions of their projected 2D points with respect to the tracked bounding boxes. The rationale is similar to the depth filter in SVO~\cite{Forster2014ICRA}, but we adapt it to bounding boxes and use a set of 3D points to estimate a target occupancy based on the previous work ~\cite{chen2020localization}. 

\subsubsection*{1) Camera Model:}
\hspace{2mm}We derive the back-projection of image points, e.g. four bounding box corners, to rays on special Euclidean group based on the models in Hartley and Zisserman's book \cite{hartley2003multiple}. A general projective camera maps a point in space $\mathbf{L}=[x, y, z, 1]^T$ to an image point $\mathbf{l}=[u, v, 1]^T$ according to the mapping $\mathbf{l} = \mathbf{P}\mathbf{L}$. $\mathbf{P}$ is the projection matrix, and $\mathbf{P}=[K|\mathbf{0}]T_c^w$, where $K_{3\times3}$ is the camera intrinsic matrix, and $T_c^w\in SE(3)$ is the transformation matrix from world frame to camera frame. We can back-project a ray from an image point,
\begin{equation}
\begin{aligned}
    \mathbf{r}_c = S(\mu K^{-1}\mathbf{l}) + \mathbf{b} \\
    \mathbf{r}_w = T_w^c \mathbf{r}_c\hspace{12.5mm}
\end{aligned}
\vspace{-1mm}
\end{equation}
where $\mathbf{r}_c$ is a homogeneous 3D point representing a ray in camera frame, $\mathbf{r}_w$ is the corresponding ray in world frame, $\mu$ is a non-negative scalar, $S_{4\times3} = \begin{bmatrix}
I_{3\times3} \\
\mathbf{0}
\end{bmatrix}$, $\mathbf{b} = [0, 0, 0, 1]^T$, and $T_w^c\in SE(3)$ is the transformation matrix from camera frame to world frame. $S$ and $\mathbf{b}$ convert Euclidean coordinates to Homogeneous coordinates. A ray in world frame is represented by a set of 3D points $\eta=\{\mathbf{r}_w| \mu \geq0 \}$. E.g. when $\mu = 0$, $\mathbf{r}_w$ is the position of camera origin in world frame.

\subsubsection*{2) 3D Points Generation: }
\hspace{2mm}We enlarge the tracked bounding boxes to compensate for errors in the UAV pose estimation. A four-face polyhedral cone, $\{\eta_{w1}, \eta_{w2}, \eta_{w3}, \eta_{w4}\}$, is generated by four 3D rays back-projected from four corners of the enlarged bounding boxed, $\{\mathbf{l}_1, \mathbf{l}_2, \mathbf{l}_3, \mathbf{l}_4\}$. To randomly initialize a 3D point in the cone, we adopt the following steps.
\begin{itemize}
\item randomly generate a direction vector from a convex combination of rays back-projected from four enlarged bounding box corners:
\begin{equation}
    \mathbf{v} = \alpha_1 K^{-1} \mathbf{l}_1 + \alpha_2 K^{-1} \mathbf{l}_2 + \alpha_3 K^{-1} \mathbf{l}_3 + \alpha_4 K^{-1} \mathbf{l}_4
\end{equation}
s.t. 
$\alpha_1, \alpha_2, \alpha_3, \alpha_4 \sim \mathcal{U}$ and 
$\alpha_1 + \alpha_2 + \alpha_3 + \alpha_4 = 1$, \\
where $\mathcal{U}$ is a standard uniform distribution.
\item
randomly scale the direction vector: $\mathbf{\Bar{L}} = \delta\mathbf{v}$, where $\delta$ is a random depth. $\mathbf{\Bar{L}}$ is a Homogeneous 3D point in the cone in camera frame.
\item
convert the 3D point to world frame: $\mathbf{L}=T_w^c(S\mathbf{\Bar{L}} + \mathbf{b})$.
\vspace{-2pt}
\end{itemize}
$\mathbf{L}$ is one randomly generated point within a cone in world frame. We generate $m$ number of 3D points for a tracked bounding box in the matrix form as presented in Algorithm 1.

{\small 
\begin{table} 
\vspace{6pt}
\resizebox{\columnwidth}{!}{%
\begin{tabular}{l}
\hline
\textbf{Algorithm 1} 3D Points Generation \hspace{20mm} \\ 
\hline
\begin{tabular}[c]{@{}l@{}}
\textbf{input}: $\{\mathbf{l}_1, \mathbf{l}_2, \mathbf{l}_3, \mathbf{l}_4\}$, $m$, $T_w^c$ \\ 
\textbf{output}: $P_w$ \\
\textbf{parameter}: $K$, $d_m$
\end{tabular}\\ 
\hline
1: \hspace{0mm} $A = \mathbb{U}(4, m)$  \hspace{25mm} 6:  $\Bar{\delta}=Diag(\delta)$ \\
2: \hspace{0mm} $B = Diag(|A|_m)$  \hspace{21mm} 7:  $\Tilde{P}_c = C\Bar{A}\Bar{\delta}$ \\
3: \hspace{0mm} $\Bar{A} = AB^{-1}$  \hspace{27.5mm} 8:  $P_c = S\Tilde{P}_c + \mathbf{b}$ \\
4: \hspace{0mm} $C = [K^{-1}\mathbf{l}_1, K^{-1}\mathbf{l}_2, K^{-1}\mathbf{l}_3, K^{-1}\mathbf{l}_4]$  \hspace{1.25mm} 9:  $P_w = T_w^c P_c$ \\
5: \hspace{0mm} $\delta = d_m\mathbb{U}(1,m)$  \\

\hline
Comments: \\
1). $m$ is the number of 3D points to be generated, e.g. $m=1000$.\\ 
2). $\mathbb{U}(x, y)$ is a standard uniform sampler returning a \\$x\times y$ matrix. \\
3). $Diag(x)$ creates a diagonal matrix from a vector $x$. \\
4). $|\cdot|_m$ returns a vector of Manhattan distances of all columns. \\
5). $d_m$ is the maximum depth of 3D points along $z$ axis \\in camera frame. \\
\hline
\end{tabular}}
\vspace*{-10pt}
\end{table}
}

\subsubsection*{3) Data Association:} 
\hspace{2mm}In assigning a tracked bounding box to an existing 3D points target, each target projects their 3D points onto the current image frame. The assignment cost metric is the number of 2D projection points inside the bounding box. The assignment is solved optimally using Hungarian algorithm. Additionally, a minimum number of 2D projection points $N_{pts}$ is imposed to reject assignment, which triggers new 3D points generation. In this study, we set $N_{pts}=m/10$, where $m$ is the total number of 3D points for a target. If a tracked bounding box is assigned to a target that has already been mapped, the remaining processes of tracking and mapping are discarded. 

\subsubsection*{4) 3D Points Update:}
\hspace{2mm}Each 3D points target maintains the UAV pose at the latest time when the target is generated or updated. Once a tracked bounding box is associated with a target, a keyframe is selected when the difference between the current UAV pose and the last keyframe pose maintained in the target is greater than a threshold. 

When a new keyframe is selected, we resample the 3D points of the associated target based on the positions between the target's 2D projection points and the bounding box in the new image frame (Algorithm 2). The distribution $f(\mathbf{p}|\mathbf{x})$ assesses the weights (importance) of 2D projection points $\mathbf{p}$ on an image given a bounding box $\mathbf{x}$, and it is formed by a combination of a Gaussian distribution and a uniform distribution:
\begin{equation}
    f(\mathbf{p}|\mathbf{x}) = w_1 h(\mathbf{p}|\mathbf{x}) + w_2 g(\mathbf{p}|\mathbf{x})
\end{equation}
where $h(\mathbf{p}|\mathbf{x})$ is distribution density function of a 2D Gaussian distribution $\mathcal{N}(\mathbf{c_x}, \Sigma_{\mathbf{x}})$, and $g(\mathbf{p}|\mathbf{x})$ is distribution density function of a 2D uniform distribution $\mathcal{U}$ among the bounding box $\mathbf{x}$. Parameters $w_1$ and $w_2$ are the weights of the two distributions. The 2D Gaussian distribution $\mathcal{N}(\mathbf{c_x}, \Sigma_{\mathbf{x}})$ is on the center of the bounding box, and \\
$\mathbf{c_x}=(\frac{u_{min} + u_{max}}{2}, \frac{v_{min} + v_{max}}{2})$, $\Sigma_{\mathbf{x}} = \begin{bmatrix}
(\frac{u_{max} - u_{min}}{2})^2 & 0 \\
0 & (\frac{v_{max} - v_{min}}{2})^2
\end{bmatrix}$, \\
where $(u_{min}, v_{min}, u_{max}, v_{max})$ are the bounding box coordinates. 
The 3D points $P_w$ in Algorithm 2 are resampled according to their weights from the joint distribution by importance sampling.

{\small 
\begin{table} 

\vspace{6pt}
\resizebox{0.95\columnwidth}{!}{%
\begin{tabular}{l}
\hline
\textbf{Algorithm 2} 3D Points Update \hspace{32mm} \\ 
\hline
\begin{tabular}[c]{@{}l@{}}
\textbf{input}: $P_w$, $\mathbf{x}$, $T_c^w$\\ 
\textbf{output}: $P_w'$ \\
\textbf{parameter}: $\sigma$, $m$, $K$
\end{tabular}\\ 
\hline
1. \hspace{5mm} $\mathbf{a} = \sqrt{\sigma}\mathbb{N}(m, 3)$ \\
2. \hspace{5mm} $\Bar{\mathbf{a}} = S\mathbf{a} + \mathbf{b}$ \\
3. \hspace{5mm} $\mathbf{p} = [K|\mathbf{0}]T_c^w (P_w + \Bar{\mathbf{a}})$ \\
4. \hspace{5mm} $\mathbf{i} = f(\mathbf{p}|\mathbf{x})$ \\
5. \hspace{5mm} $P_w' = Importance\_Sampling(P_w, \mathbf{i})$  \\
\hline
Comments: \\
1). $\mathbb{N}(x, y)$ is an isotropic standard multi-variate Gaussian \\sampler returning a $x\times y$ matrix. \\
2). $\sqrt{\sigma}$ is the standard variance of update noise. \\
3). $f(\mathbf{p}|\mathbf{x})$ is a two-variable distribution composed of a \\Gaussian distribution and a uniform distribution. \\
\hline
\end{tabular}}
\vspace*{-10pt}
\end{table}
}

\subsubsection*{5) Properties:}
\hspace{2mm}Two statistical metrics are used to characterize 3D points and determine a target converging state. We use differential entropy (DE), which indicates the compactness of a set of 3D points, to determine if a target is converging. To compute DE, we approximate the 3D points set with a three-dimensional Gaussian distribution, whose mean and covariance are approximated by the mean and covariance of the 3D points. The DE is computed from the following formula for multivariate Gaussian distribution:
\begin{equation}
h=\frac{k}{2} + \frac{k}{2}\ln(2\pi) + \frac{1}{2}\ln(|\Sigma|) 
\end{equation}
where $k=3$ is the dimensionality of the state vector. When $h$ is greater than a threshold $h_c$, we change the target state to converging.   

When a set of 3D points is updated, we want to measure the amount of information gain from the update. One metric that captures relative entropy is Kullback–Leibler divergence (KLD) of updated 3D points distribution with respect to previous 3D points distribution. We compute the KLD following the formula for multivariate Gaussian distribution:
\begin{equation}
\begin{aligned} \scriptstyle
    D_{KL}(\mathcal{N}_0||\mathcal{N}_1) = \frac{1}{2}(tr(\Pi_1^{-1}\Pi_0) + (\tau_1 - \tau_0)^T\Pi_1^{-1}(\tau_1 - \tau_0) - k + \ln(\frac{|\Pi_1|}{|\Pi_0|})) 
\end{aligned}
\end{equation}
where $k=3$ is still the dimensionality of the state vector, $\mathcal{N}_0(\tau_0, \Pi_0)$ is the estimated Gaussian distribution of the updated 3D points, $\mathcal{N}_1(\tau_1, \Pi_1)$ is the estimated Gaussian distribution of the previous 3D points, and $tr(\cdot)$ is a matrix trace function. Because we have no prior knowledge of targets, KLD is useful to determine if a target is converged by measuring the similarity of two 3D points distributions. In practice, KLD of a target has to be smaller than a threshold for $N_{KLD}$ continuous updates to be determined as converged.

\subsubsection*{6) Target State Update:}
\hspace{2mm}The 3D points tracker processes regular tracking as described above once receiving a tracked bounding box. When there is no message from the bounding box tracker, the 3D points tracking process is called by a timer interrupt. However, the timer interrupt is only in charge of target deregistration, which is slightly different from the regular tracking process. A target is permanently registered once mapped. Otherwise, if a target is under tracking (the target is not converged yet), it is deregistered when it has not been continuously updated from any tracked bounding box for $T_{pts\_missing}$ tracking iterations including regular tracking processes or timer interrupts. We introduce such a grace period to deal with false positive and missing detections that are not cut out by the bounding box tracker. E.g. a 3D points target generated from false positive detections cannot have enough updates to get converged because the false positive detections are eliminated from some different camera perspectives. Similarly, such grace period allows a certain degree of missing detections as a PBR is re-detected by the neural network.

\subsection{Target Mapping}
Once a target is converged, we first examine if the target has already been mapped by checking the distances between this target center and other mapped target centers. If all distances are greater than a threshold, an RGBD SLAM \cite{labbe2019rtab} is deployed for target mapping. When the target mapping is done, the target state is changed to mapped, and its corresponding 3D points are replaced by a downsampled point cloud from the mapping. The original dense point cloud of the target is saved to a hard disk. 

\section{Motion Subsystem}
We start by carrying out a lawn-mower search pattern in a given survey region from a high elevation. Once the state of a 3D points target from the 3D points tracker is assessed as converging, the UAV deploys a circular motion at the same lawn-mower search elevation. The purpose of this circular motion is to keep filtering the 3D points target and ultimately estimate a b-cylinder of a PBR from the converged 3D points. Based on the b-cylinder, the UAV conducts a path planning to closely and completely map the target with SLAM. After target mapping, the UAV resumes the lawn-mower search pattern to find the next target. The motion state diagram is illustrated in Fig.~\ref{motion_state}.
 
We adopt only one control system for all motion states including lawn-mower search, b-cylinder estimation motion, and mapping motion. The control system is illustrated in Fig.~\ref{control_system}. The motion states switch by altering path generators. In the target search state, the UAV path is a hard-coded lawn-mower pattern. For the b-cylinder estimation motion and the mapping motion, the path generators are based on view planning methods, which are introduced in the following subsections. The trajectory generator takes keypoints from the path generator and generates a trajectory that minimizes the snap of 10th-degree polynomials \cite{richter2016polynomial}. The geometric controller takes a desired robot's position and yaw heading and outputs desired robot's body rate and thrusts \cite{jaeyoung_lim_2019_2619313, lee2010geometric}. The attitude controller customized with the UAV hardware configurations by the flight controller produces desired motor speeds for UAV propellers.

Because the geometric controller is only concerned with desired position and yaw, we add maximum acceleration and maximum velocity constraints in the trajectory generator such that UAV's pose can be approximated to hovering. That is, UAV's roll and pitch angles are neglected in the view plannings. Additionally, because the distance $d_1$ between a PBR center $C_e$ and the camera is much larger than the distance $d_2$ between the camera and the UAV ($d_1:d_2>10:1$), we do not differentiate camera position or UAV position in the view plannings.

\subsection{Bounding Cylinder Estimation Motion}
Because camera perspectives from lawn-mower search alone cannot always precisely estimate PBR's occupancy, b-cylinder estimation motion is activated when a 3D points target is assessed as converging during lawn-mower search. The path generator then produces a set of camera perspectives to reduce uncertainty in the target occupancy.

\begin{figure} 
    \centering
  \subfloat[View planning for b-cylinder estimation\label{estimation_motion}]{%
       \includegraphics[width=0.5\linewidth]{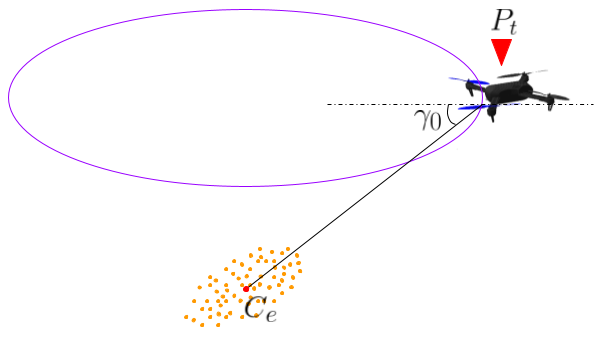}}
    \hfill
  \subfloat[View planning for target mapping  \label{mapping_motion}]{%
        \includegraphics[width=0.45\linewidth]{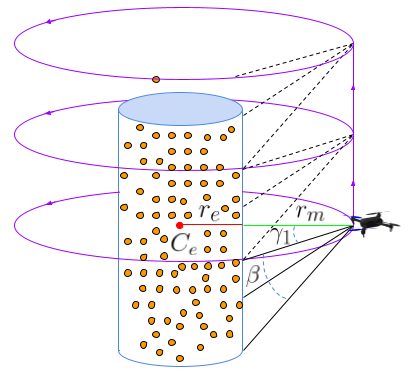}}
  \caption{(a) Large ellipse on the top is the circular path. $C_e$ is the geometric center of the 3D points. $\gamma_0$ is the angle between $\overline{C_eP_t}$ and horizontal plane. $P_t$ is the start position on the circular path. (b) Three illustrated circles are the generated path for mapping in this case. Small yellow dots represent converged 3D points of a target. Blue cylinder is b-cylinder from the 3D points. $C_e$ is the center of the 3D points (mean Euclidean distance). $r_e$ is the radius of the cylinder. $r_m$ is the distance to be kept between the UAV and the cylinder surface. $\beta$ is the vertical scanning field of view. $\gamma_n$ is the angle between upper scanning ray and horizon plane.}
  \label{fig:localization_mapping} 
\end{figure}

We approach the view planning problem by first introducing geometric constraints and then finding a feasible solution. Given the angle $\gamma$ between the camera $z$ axis and the horizontal plane, there are four variables in UAV's pose to be determined, $(x, y, z, yaw)$. We first determine $z$ by constraining the UAV positions on the lawn-mower search plane, which is collision-free and also parallel to $XY$ plane in world frame. Because the view planning depends on 3D points, which might have large uncertainty before being refined by the b-cylinder estimation motion, setting the camera $z$ axis to pass through the geometric center of 3D points $C_e$ can keep the target projection in the center of image frame as much as possible. However, this is problematic when the camera is mounted vertically straight down. In practice, we relax this constraint and set the camera $z$ axis to pass through a vertical line where the 3D points center $C_e$ lies such that the angle between $C_e$ and horizontal plane is $\gamma_0$, i.e. $\gamma_0 = 45\degree$ (Fig.~\ref{estimation_motion}). The UAV's heading $yaw$ is correspondingly determined. From the above geometric constraints, a circular path is constructed as the path of the b-cylinder estimation motion, where the remaining $(x, y)$ are also determined. From the current UAV pose where a 3D points target is assessed as converging, the next keypoint from the path generator is the closest point on the circle $\mathbf{C}_l$. Then the UAV follows the circle curve $\mathbf{C}_l$ to keep filtering 3D points. 

When 3D points are assessed as converged, we fit the 3D points with a vertical cylinder, which is the intersection of two other vertical cylinders $H=H_1 \cap H_2$. $H_1$ is the smallest vertical cylinder that includes all 3D points and has the axis traversing the center of the 3D points $C_e$. $H_2$ has the cylinder center at $C_e$; its height is the six sigma value of the 3D points along $z$ axis; its radius is the largest three sigma value along $x$ or $y$ axis. The b-cylinder is used for view planning of target mapping. 

\subsection{Mapping Motion}
The path generator for target mapping should produce enough camera perspectives to densely map the complete visible body of a target. We deploy a circular path for the mapping (Fig.~\ref{mapping_motion}). The circles lie on horizontal planes that are parallel to $XY$ plane in world frame. The centers of the circles are on the axis of the b-cylinder. We set the circle radius $r_c =r_e+r_m$ to keep some distance between the UAV and the b-cylinder. The spacing and number of the circles are determined by $r_c$, the upper scanning angle $\gamma_1$, and vertical scanning field of view $\beta$ (Fig.~\ref{fig:localization_mapping}) such that the cylinder surface excluding the bottom will be fully covered by the camera scanning views. We have $\beta$ smaller than the camera vertical field of view so that there are vertical overlaps between circular scans. The first circle is placed at the bottom where the cylinder bottom can be just covered by the lower scanning ray. The other circles are repeatedly arranged upward until the center of cylinder top is covered by the upper scanning ray.

\subsection{Motion State Update}
As the motion state diagram shown in Fig.~\ref{motion_state}, if 3D points have not been converged after one circular movement of b-cylinder estimation, the system fails to verify the 3D points target and thus deregisters the 3D points target and resumes the lawn-mower search. When a target gets deregistered during the b-cylinder estimation motion due to missing detections, the motion state also switches back to the lawn-mower search. 

Because the transition of the motion states is triggered by target states from 3D points tracking, when there are multiple targets with different states, we rely on a priority table to decide the motion state. The priorities descend from resuming lawn-mower search, to mapping motion, to b-cylinder estimation motion. When a UAV is already on b-cylinder estimation motion state for a target, this target has higher priority than other targets that are also assessed as converging. In other scenarios, when there are multiple targets with the same states, the targets are processed as a queue data structure.  Additionally, we hash target states for each target for the efficiency and convenience of inquiry. 

\begin{figure} 
    \centering
  \subfloat[UAV trace of Experiment I.\label{granite_dell_motion}]{%
       \includegraphics[width=0.47\linewidth]{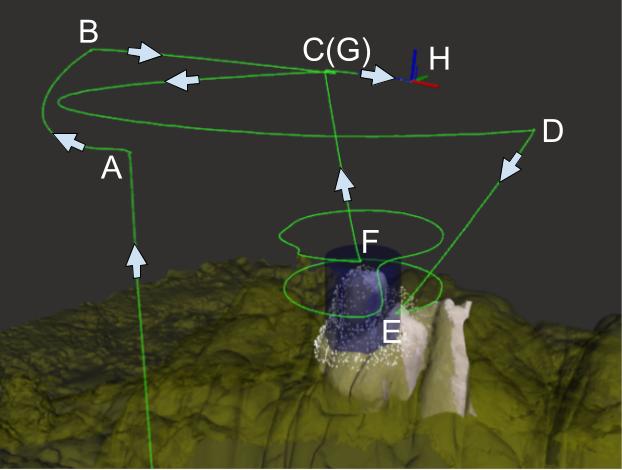}}
    \hfill
  \subfloat[UAV trace of Experiment II. \label{blender_motion}]{%
        \includegraphics[width=0.52\linewidth]{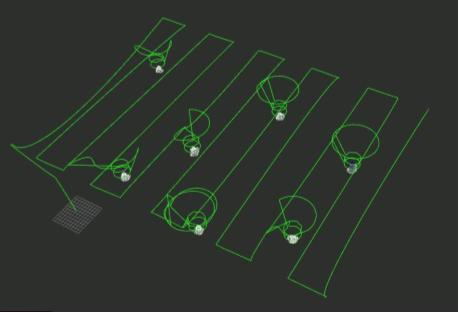}}
  \caption{UAV traces. (a) $\overline{ABC}$ is lawn-mower search trace. $\overline{CD}$ is b-cylinder estimation motion trace, and target is assessed as converging at $C$ and as converged at $D$. $\overline{EF}$ is the target mapping trace.  $\overline{GH}$ is resumed lawn-mower search trace.}
  \label{fig:exp_motion} 
\end{figure}

\begin{figure} 
    \centering
  \subfloat[Gazebo world for Experiment II \label{blender_gazebo}]{%
       \includegraphics[width=0.475\linewidth]{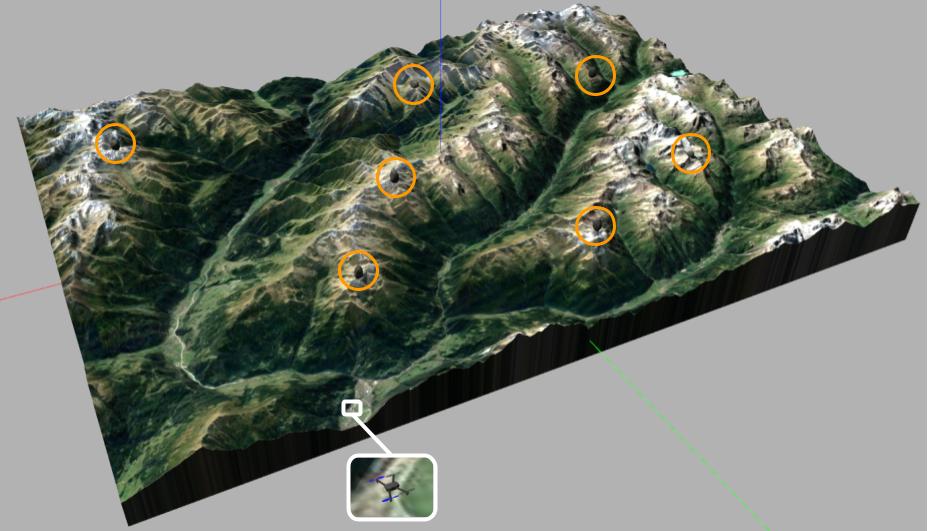}}
    \hfill
  \subfloat[Mapping results \label{blender_result}]{%
        \includegraphics[width=0.515\linewidth]{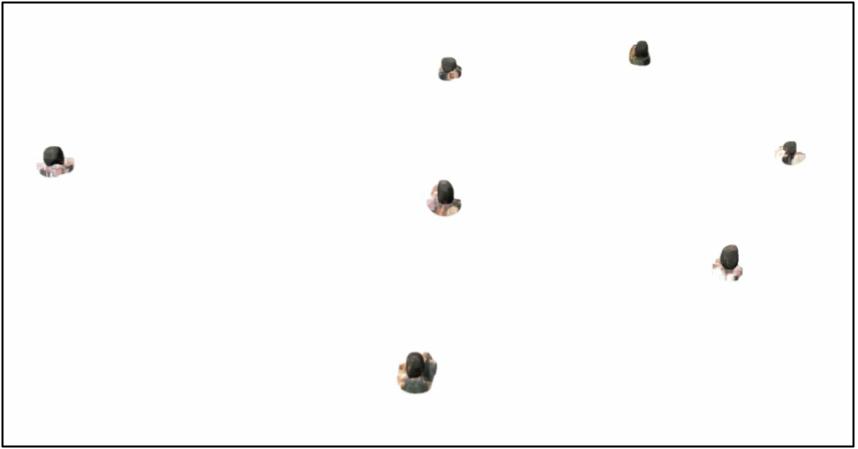}}
  \vspace{-10pt}
  \caption{Gazebo world and mapping results from Experiment II.}
  \label{fig:blender_terrain} 
  \vspace{-2pt}
\end{figure}

\section{Experiment}
We evaluate the target-oriented mapping system with two experiments in Gazebo simulation using a gaming laptop, Dell G7: Intel Core i7-8750H, 16GB RAM, and Nvidia GeForce GTX 1060 Max-Q, which has similar computing capacity with Intel NUC i7 plus Nvidia TX2. In both experiments, we use a 3DR Iris with an RGBD sensor ($\gamma = 60\degree$) operated by PX4 SITL and MAVROS. The UAV pose estimation is maintained by an EKF fusing data from IMU in the flight controller and a GPS module. Sensor noises of GPS, IMU, and magnetometer are also simulated in Gazebo. A tiny YOLO v2 is initialized from the darknet53 ~\cite{redmon2017yolo9000} and finely tuned with the images collected in the Gazebo worlds, and the real-time inference is deployed using YOLO ROS~\cite{bjelonicYolo2018}. We apply RTAB-MAP~\cite{labbe2019rtab} for RGBD mapping, and the mapping service is only activated during the mapping motion. The vertical scanning field of view $\beta$ is $40\degree$. For the control system, the maximum velocity is $1m/s$ and the maximum acceleration is $1m/s^2$. 

The first experiment is conducted to demonstrate the target-oriented mapping pipeline for a field reconstructed from real data. We build the mesh model with one PBR (Fig.~\ref{fig:granite_dell} (bottom left)) using SfM with UAV imagery that was manually collected at Granite Dells, Prescott AZ, 2019. The UAV trace with different motion states is shown in Fig.~\ref{granite_dell_motion}. As the mapping result presented in Fig.~\ref{fig:granite_dell} (right), the presented system has mapped the target and also provided access to the complete visible surface features of the PBR including the ground basal contact, which is crucial to calculate the PBR's fragility for earthquake studies. 

{\small
\begin{table}[h]
\centering
\vspace*{5pt}
\captionsetup{font=scriptsize}
\caption{Target-oriented Mapping System Performance}
\begin{adjustbox}{width=0.95\columnwidth}
\begin{tabular}{|c|c|c|c|c|c|}
\hline
\multirow{2}{*}{}  & worst     & \multicolumn{4}{c|}{all}       \\ \cline{2-6} 
                   & detection & generation & converging & converged & mapped \\ \hline
precision          & 88.7\%       & 84.6\%        & 88.9\%        & 100\%       & 100\%    \\ \hline
recall             & 78.3\%       & 100\%        & 100\%        & 87.5\%       & 100\%    \\ \hline
\end{tabular}
\end{adjustbox}
\label{table:metrics}
\vspace*{-5pt}
\end{table}}

The purpose of the second experiment is to examine the system with multiple sparsely distributed targets. The 3D terrain model (Fig.~\ref{blender_gazebo}) is built in Blender~\cite{blender} with texture from satellite imagery and digital elevation model from OpenTopography~\cite{amatulli2020geomorpho90m}. We import seven synthesized PBRs to the terrain model as our mapping targets. The UAV trace is shown in Fig.~\ref{blender_motion}, and the mapping results are displayed in Fig.~\ref{blender_result}.

We show the system robustness by measuring the performances of key components in Experiment II (Table.~\ref{table:metrics}). As false positive and missing detections are the major cause of the problems in the following components, we inspect the worst neural network detection performance (IoU>0.5) in one of the seven PBR mappings, where each begins from the end of a previous PBR mapping to the converged assessment of a current PBR. The other components are measured for the entire seven PBR mappings based on the target states presented from the system and the desired target states without considering mission duration. For the false positives and false negatives of converging, we only consider the cases that are caused by false detections from the neural network. We neglect the cases where there are not enough detections due to the bounding box is too close to the image edges, because they are eventually detected from other search perspectives. This experiment shows the robustness of the system as all seven PBRs are successfully mapped, especially with regard to the existence of false positives and false negatives in other components.

\section{Conclusion and Future Work}
We present a target-oriented system for UAV to map fragile geologic features such as PBRs, whose geometric fragility parameters can provide valuable information on earthquake processes and landscape development. Such a system provides access to autonomous PBR mapping, which was lacking in field geology but will be essential to geologic model assessment.  

Our ultimate goal is to deploy this target-oriented mapping system to actual boulder fields and to assess the quality of 3D mapping. Additionally, 3D semantic segmentation should be applied to extract 3D points of PBRs from their contact terrains, which leads to PBR surface reconstruction. We have implemented the circular motion for target mapping, which is based on an assumption that there are no obstacles during the mapping. This can be future improved by framing the mapping as a probabilistic exploration problem. Next-best-view or frontier-based exploration algorithms~\cite{bircher2016receding, song2018surface} can be used to generate an obstacle-free path and accurately determine occupancy within the bounding cylinder. 

\section*{ACKNOWLEDGEMENTS}
This work was supported in part by Southern California Earthquake Center (SCEC) award 20129, National Science Foundation award CNS-1521617, National Aeronautics and Space Administration STTR award 19-1-T4.01-2855, and a gift from Pacific Gas and Electric Company.

{\small
\printbibliography}

\end{document}